\relax

\documentclass[letterpaper]{article} 
\usepackage{aaai18-custom}  
\usepackage{times}  
\usepackage{helvet}  
\usepackage{courier}  
\usepackage{url}  
\usepackage{graphicx}  
\usepackage[utf8]{inputenc} 
\usepackage{tabularx}

\usepackage[usenames,svgnames]{xcolor}

\usepackage{booktabs} 

\title{Generating Paths with WFC}

\author{Hugo Scurti\\
    McGill University\\
    Montréal, Québec, Canada\\
    \texttt{hugo.scurti@mail.mcgill.ca}
    \And
    Clark Verbrugge\\
    McGill University\\
    Montréal, Québec, Canada\\
    \texttt{clump@cs.mcgill.ca}
}

\begin{document}
\maketitle

\begin{abstract}
Motion plans are often randomly generated for minor game NPCs. Repetitive or regular movements, however, require non-trivial programming effort and/or integration with a pathing system.  We here describe an example-based approach to path generation that requires little or no additional programming effort.  Our work modifies the \textit{Wave Function Collapse} (WFC) algorithm, adapting it to produce pathing plans similar to an input sketch.  We show how simple sketch modifications control path characteristics, and demonstrate feasibility through a usable Unity implementation.
\end{abstract}

\section{Introduction}

Generating randomized paths for \textit{Non-Player Characters} (NPCs) is a common task in game design.  Simple random walks,
however, are unsatisfying, in tending to unrealistic, jagged, fractal-like trajectories with erratic
space coverage~\cite{lawler}.  Better results are possible by making use of higher-level navigation
models, such as a waypoint graphs or flow models, but it is in general not an easy task to do for
non-programmers, requiring substantial algorithmic development and significant programming effort to
ensure simple, frequently desired properties, such as avoiding self-intersection, forming closed loops, and encircling
obstacles.

In this work we describe an example-based approach to path generation with a simple, flexible interface. Our design is based on the \textit{Wave Function Collapse} (WFC) algorithm~\cite{GuminWFC}, an example-based approach to texture generation.  We modify and augment WFC to form a workflow that takes a representative path design and game level as input, and produces a usable set of paths respecting the given path properties and level (obstacle) constraints.  Simple modifications to the input path design can be done with any image editor, and can then be used to control properties of the resulting paths. Our approach is realized in the Unity framework, and we show how the algorithm scales to significant, practical level sizes.  For easy use and experimentation, the tool is able to read in actual game levels expressed in ASCII form, as from the \textit{Moving AI} benchmark suite \cite{sturtevant2012benchmarks}.

Contributions of our work are as follows:
\begin{itemize}
\item We define modifications to the basic WFC algorithm to allow for small, generic images to be used as   general models of path design.  Basic, easily modified properties of the image then control useful properties of the path output.  Our approach includes non-trivial changes necessary to accommodate the densely constrained output context of typical game levels.
\item A complete realization in Unity is shown and made available
\footnote{https://github.com/hugoscurti/path-wfc}
, including a full workflow that reads in game levels and generates abstract path data structures.  This work includes post-processing steps to filter and smooth the output paths.
\end{itemize}

\section{WFC Background}
Our approach builds heavily on the \textit{Wave Function Collapse} algorithm developed by \citeauthor{GuminWFC} \shortcite{GuminWFC}, more precisely the overlapping version of the algorithm. This algorithm builds random output textures based on smaller input textures by looking at overlapping patterns in the input texture sample. Understanding this algorithm is critical to understanding our modifications, so we summarize the main steps below.  Further detail can be found in the code itself.
\begin{enumerate}
    \item Extract all possible unique $n \times n$ patterns from a discretized input image; the frequency of each unique pattern in the input is also later used as a sampling distribution. \label{wfc:stepA}
    
    \item Compute all the possible overlap combinations between each pattern. The (im)possibility of overlap between patterns results in the main set of constraints used to build an output image.\label{wfc:stepB}
    
    \item Each discretized tile in the output sample maintains a list of legal patterns that could define the pixel value(s) in that tile. Absent any constraints, such as at the start of the algorithm, every pattern is legal in any tile.\label{wfc:stepC}
    
    \item Choose a tile in the output sample and commit the value based on one of the legal patterns that cover the tile.  This choice follows an entropy calculation that looks for the tile that has the fewest available patterns, and thus is most constrained. The frequency distribution of basic patterns computed in step~\ref{wfc:stepA} is used to bias selection of the pattern. This is referred to as the \textbf{observe} step of the algorithm.\label{wfc:stepD}
    
    \item 
    The overlap combinations computed in step~\ref{wfc:stepB} are used to filter out any now illegal patterns covering adjacent positions, and this is applied recursively throughout the output image. This is referred to as the \textbf{propagate} step.\label{wfc:stepE}
    
    \item Repeat steps~\ref{wfc:stepD} and~\ref{wfc:stepE}, observing and propagating until no further progress can be made. At that point either all tiles are fixed to one pattern (the algorithm succeeded), or at some point a tile has no more patterns available (the algorithm failed). In the latter case the algorithm is restarted with a different random seed.\label{wfc:stepF}
\end{enumerate}

The original implementation adds a few helpful options for texture generation, such as the ability to set boundary patterns, wrap periodically, and extend the range of available patterns by adding rotations and reflections of patterns sampled from the input.

\section{WFC for Path Generation}

We aim to use WFC to enable example-based path generation.  A user supplies a single path sketch, showing an example path, and this is applied to an output level, filling the level with paths that have characteristics similar to the original example.

Unfortunately, while the basic WFC algorithm excels at creating randomized images respecting the constraints of immediate pattern adjacency, it has difficulty guaranteeing non-local properties.  For path generation we require a mixture of local properties, such as angle of movement and proximity to obstacles, and properties that are highly non-local, such as control over relative shape, the ability to form closed curves, and encircle/contain other features. Our output images are also existing game levels, already densely constrained by obstacles, complicating the creation of initial constraints and making WFC's core trial-and-error approach highly inefficient, or even impossible.  Below we describe 3 main techniques we used to modify WFC to better handling pathing.

\subsection{Output constraints and pre-processing}
The original WFC algorithm lets users fix specific tiles in the output to enforce initial output constraints.  This builds on the basic algorithm process, manually choosing/reducing the set of legal patterns at a specific location and propagating the resulting constraints. For a large map, filled with obstacles, this would be extremely tedious, and some initial pre-processing is thus necessary to dynamically fix tiles in the output to guarantee the result continues to faithfully represent the original game level.

The process is essentially a batched, automated version of the basic manual fixing.  We filter the input level representation, constructing an initial output image with tiles designed as either free-form, pathable space (represented by a white pixel) or obstacles (represented by a red pixel); path fragments (black pixels) could also be incorporated.  Interiors of obstacles are thus constrained to a single pattern based on being filled with a single, unique colour, while the patterns permitted to cover perimeters of obstacles are reduced to match the boundary edge.

\begin{figure}[htbp]
    \centering
    \includegraphics[width=2.8in]{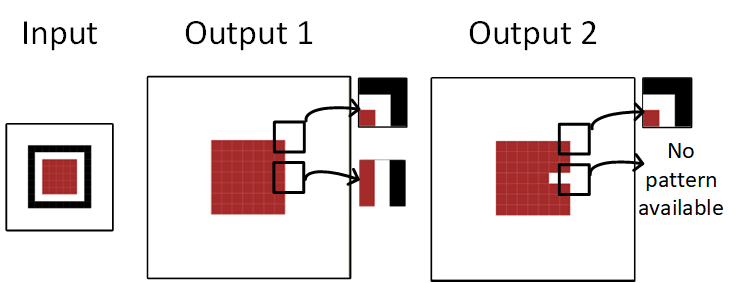}
    \caption{Different outputs making the algorithm succeed (output 1) and fail (output 2).}
    \label{fig:FixedOutputGoodBad}
\end{figure}

The process is flexible, and also provides early warning of unsatisfiable constraints.  Figure \ref{fig:FixedOutputGoodBad} shows a specific \textit{Input} path image/template, representing a path encircling an obstacle.  For \textit{Output~1} a solution is possible; a level such as \textit{Output~2}, however, fails, as the notched obstacle cannot be filled by any pattern found in the input.

\subsection{Arbitrary colours as ``stretch space''}

\begin{figure}[ht]
    \centering
    \includegraphics[width=2.8in]{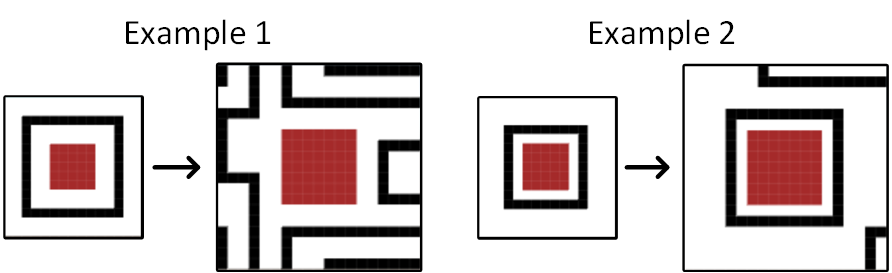}
    \caption{Execution of the algorithm without using stretch space.}
    \label{fig:NonstretchExamples}
\end{figure}

An important goal is that generated paths have at least some heuristic, qualitative similarity to the example input. Overly literal interpretation of the input image, however, tends to generate paths that are either too strict or under-constrained.  Figure~\ref{fig:NonstretchExamples} shows examples of trying to use an input sketch to imply paths should generally encircle the obstacle.  In example~1 the result is under-constrained.  The distance between the example path and the obstacle means patterns exist that do not contain both the path and obstacle colours.  Generated paths thus meander arbitrarily, more or less oblivious to the obstacle. Example~2 shows an attempt to solve this by putting the path close to the obstacle, but this results in only generating paths strictly one unit away from the obstacle (or from another path edge).  

We introduce the notion of \textit{stretch space} to alleviate this issue. We use a unique, designated color in the input image to represent the space between an obstacle and the path that goes around it. We make sure that one pattern can be filled entirely with this color so that the overlapping nature of the algorithm will make it overlap with itself, therefore making the space between obstacles and paths \textit{stretchable}.
This helps generated paths behave more closely to their inputs as it restricts the sets of overlapping patterns for each pattern that contains a path. 

\begin{figure}[ht]
    \centering
    \includegraphics[width=2.8in]{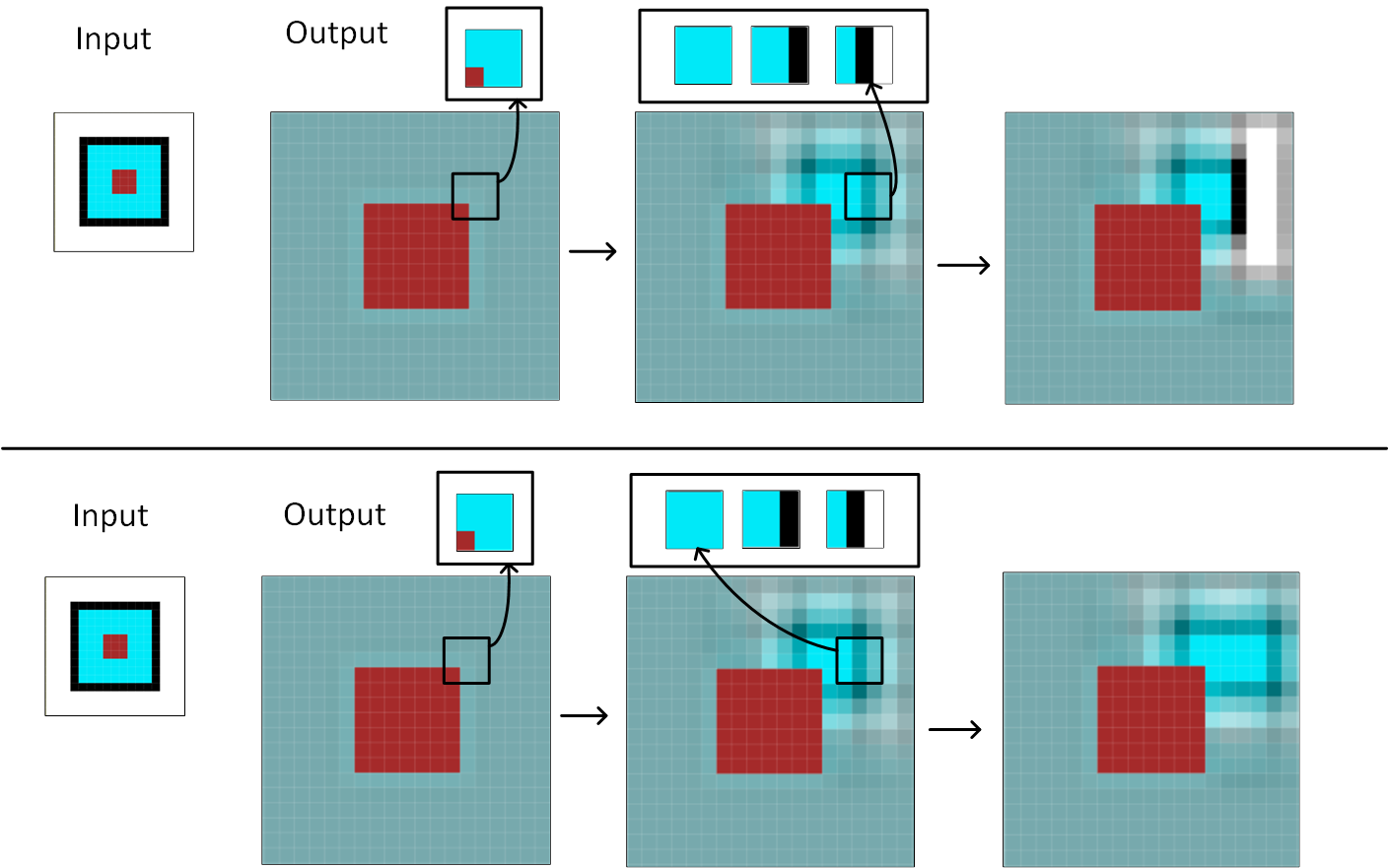}
    \caption{\begin{small}
    Example of using stretch space propagating out from an obstacle corner. On the top, the stretching is terminated by a vertical path line, defining a generated path one unit away from the obstacle.  The bottom instance shows a different, random choice, selecting empty stretch space to push the path tile away further.
    Note that similar to Gumin's implementation, the images show progress during the execution of the algorithm, averaging the color from the set of all surrounding legal patterns until it is filled by one pattern, rather than just showing blank/white pixels. This results in a blurry effect while patterns are being unveiled.
    \end{small}}
    \label{fig:StretchSpace}
\end{figure}

Figure~\ref{fig:StretchSpace} shows an example, using light blue to indicate stretch space.  The blank area of the output can then be filled with stretch space, ensuring path pixels have some relation to obstacles (or other path segments): blue patterns may be selected to extend the stretch area, eventually and necessarily reaching path/obstacle patterns that close the path.

Using stretch space allows for greater versatility and generally better correspondence with the path behaviour of the input example.  It is, however, a heuristic trade-off in permissiveness.  Patterns will typically exist with no obstacle pixels, allowing construction of (usually smaller) paths enclosing only stretch space, and thus not necessarily encircling an obstacle. We give the option to remove these as a post-processing step.

\subsection{Masks}
Input examples are meant to be simple, schematic drawings.  Outside trivial situations, these are unlikely to include all possible patterns needed to fill the output level map---complex and varied obstacle and boundary shapes in the intended level space generate output constraints that will in general not be satisfiable. 
Path generation, however, has the advantage over other WFC applications in that it is not strictly necessary to find a complete solution.  A partial solution that establishes paths can be a sufficient, even if the set of pattern constraints cannot be fully propagated to cover every tile in the output.

We thus introduce \textit{masks} to help cover tiles for which we do not have an appropriate input pattern.  We do this by automatically creating patterns from the output map. We analyze the output for all patterns that contain obstacles. Then, for each pattern, we replace white tiles around obstacles by various masks depending on their relative position to an obstacle. Tiles that are adjacent to an obstacle contain a mask that allows everything but the path color and all other tiles contain a mask that allows any color.  This design enables input-derived patterns to propagate up to obstacles, but not fail constraints due to being unable to align with the exact obstacle boundary---even if not explicitly sketched in the input example, any obstacle in the output can be dealt with and can have paths, free space and stretch space around it. 

\begin{figure}[ht]
    \includegraphics[width=3.2in]{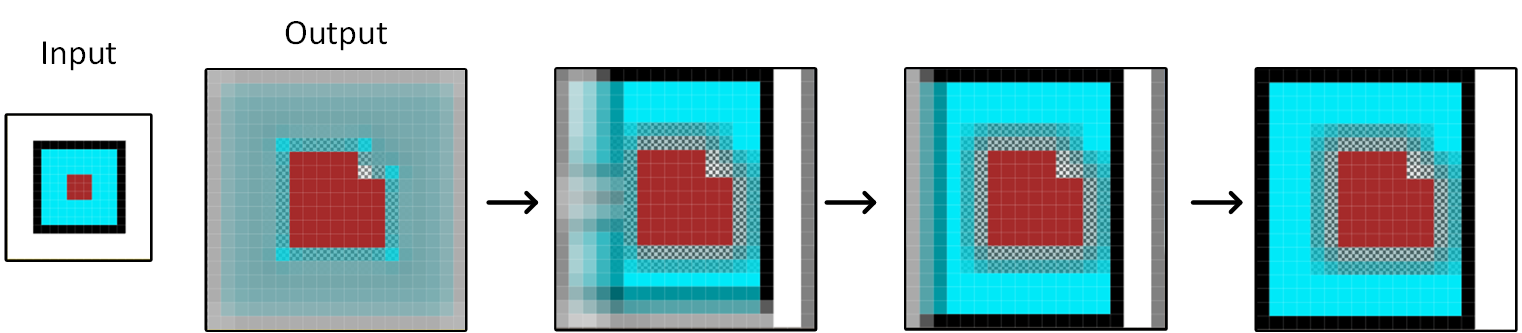}
    \caption{\begin{small} Execution of the algorithm using masks on output boundaries. Masks are represented as transparent tiles.\end{small}}
    \label{fig:MaskExecution}
\end{figure}

Figure~\ref{fig:MaskExecution} shows an example. The input example does not include a concave notch in the obstacle, and thus a naive WFC will be unable to find a solution given the limited set of patterns present in the input.  Using a mask we can effectively ignore the aberration, successfully generating a cyclic path around the obstacle even though no pattern from the input fits the upper right notch.

Masks obviously represent a trade-off between successful generation and strictly following input constraints.  Our tool design allows masks to be toggled on or off, depending on the path generation goals.  

\subsection{Post processing}
\begin{figure}[ht]
    \centering
    \includegraphics[width=2.8in]{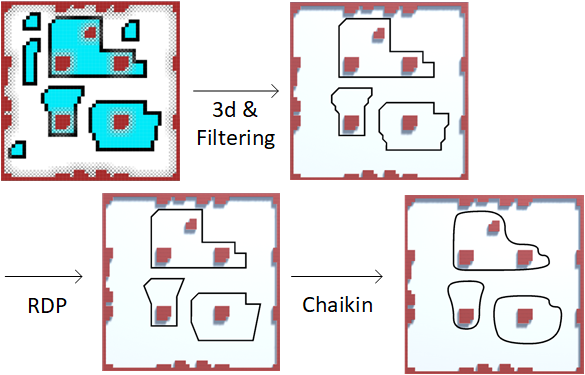}
    \caption{\begin{small} Execution of all post-processing steps in their respective order. Output map is ``arena'' from the \textit{Dragon Age: Origins} benchmark set.\end{small}}
    \label{fig:PostProcessing}
\end{figure}

The heuristic nature of our approach is augmented with some post-processing steps to filter out degenerate constructions, help visualize results, and generally better prepare paths for integration into a game environment.
Figure~\ref{fig:PostProcessing} shows an execution flow for of all processing steps. With the exception of the initial conversion, all steps mentioned below are optional and can be used independently.

\subsubsection{Conversion to usable path}
The WFC image output is converted into a waypoint format that can be used in a game environment.  We generate paths by following path tiles incrementally until a path is complete. The actual 3D model requirements are of course game dependent, but we provide a baseline conversion for easy visualization, using simple cubes for obstacles, and line renderers for paths.

\subsubsection{Path filtering}

As a result of using overlapping patterns and masks, outputs may contain smaller paths that are not necessarily appealing. We enable the option to filter out paths that are smaller than a specified length threshold.

\subsubsection{Path Simplification}

Depending on the input patterns, some paths can get complicated and have a significant amount of redundant turns or detours.  We provide a post-processing option to simplify paths, based on the \textit{Ramer-Douglas-Peucker algorithm (RDP)} \cite{RAMER1972244,DouglasPeucker} to remove redundant vertices. 
We modified the algorithm to avoid obstacle intersection when simplifying a path segment.

\subsubsection{Path Smoothing}

Since they are generated using $3 \times 3$ patterns, output paths naturally have sharp turns and corners. We thus also provide smoothing using \citeauthor{Chaikin1974}'s curve generation algorithm \cite{Chaikin1974}. 
Similar to path simplification, we modified the algorithm to make sure that we do not cut a corner if a modified path segment intersects an obstacle.

\section{Experimentation}
The precise intention of a user's input path example is of course open to multiple interpretations.  
Experimental work is thus aimed at providing proof-of-concept evidence that the input image does allow for control over the output, with a qualitatively clear correspondence of features.  We also address performance concerns, showing that the process scales to relatively large output maps typical of modern games.  

\subsection{Setup}

Our setup is based on a custom Unity implementation of our modified WFC algorithm. We use custom inputs, extracting patterns of size $3 \times 3$ as sufficient to represent useful path properties without introducing major scaling issues. Input and output 2D maps use small monochromatic sprites to represent tiles and interactively show the execution of the algorithm. When the execution succeeds, post processing steps can be applied to generate paths in a 3D representation of the output map, along with an agent that can walk through generated paths. The algorithm was used on a 64-bit Windows 10 machine with an Intel Core i5-4200U CPU running at 1.60 GHz, an Nvidia GeForce GT 750M GPU, and 8 GBs of RAM.
The algorithm was tested with maps representing game levels from \textit{Dragon Age: Origins} and \textit{Baldur's Gate II}, as extracted from the Moving AI benchmark sets \cite{sturtevant2012benchmarks}. 

\subsection{Controlling the output}
Output properties are affected by changes to the base WFC algorithm, as well as by the given input example.  Below we show representative examples to illustrate the range of results and give an indication of how the output can be controlled.

\subsubsection{Stretch space}
Use of stretch space was motivated by the desire for better correspondence with the input.  Both types of input, with and without stretch space, however, generate interesting paths that may be useful for different purposes, with the latter having a much higher degree of randomness.

\begin{figure}[ht]
    \centering
    \includegraphics[width=2.8in]{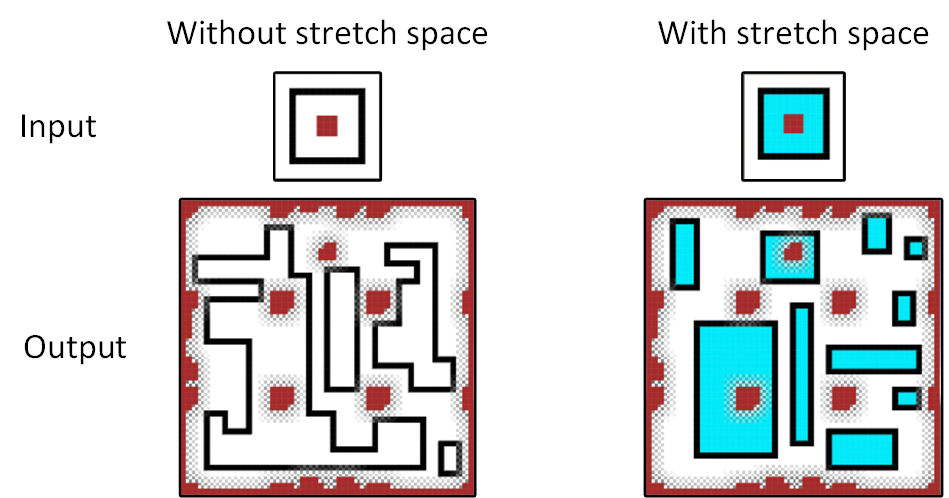}
    \caption{\begin{small}Comparison between output generated without and with stretch space. 
    \end{small}}
    \label{fig:BufferVsNonBuffer}
\end{figure}

Figure~\ref{fig:BufferVsNonBuffer} shows the effect of using stretch space or not in an otherwise identical input/output context. 
In contrast to example 1 from figure~\ref{fig:NonstretchExamples} where paths are not closed, execution without stretch space (figure~\ref{fig:BufferVsNonBuffer}, left) in a larger output bounded by obstacles results in reasonable closed routes, although more meandering than the rectangular paths generated by using stretch space (right).
The effect of stretch space is driven by the constraints between pattern overlaps, where having different colors inside and outside the cyclic path prevents corners from acting as both inside and outside turns.
This gives us two different ways of generating paths with one input, using a less constrained interpretation to generate a broader range of outputs, or a stricter view that generates results more similar to the shape of the input path.

\subsubsection{Frequency distribution}

\begin{figure}[htbp]
    \centering
    \includegraphics[width=3in]{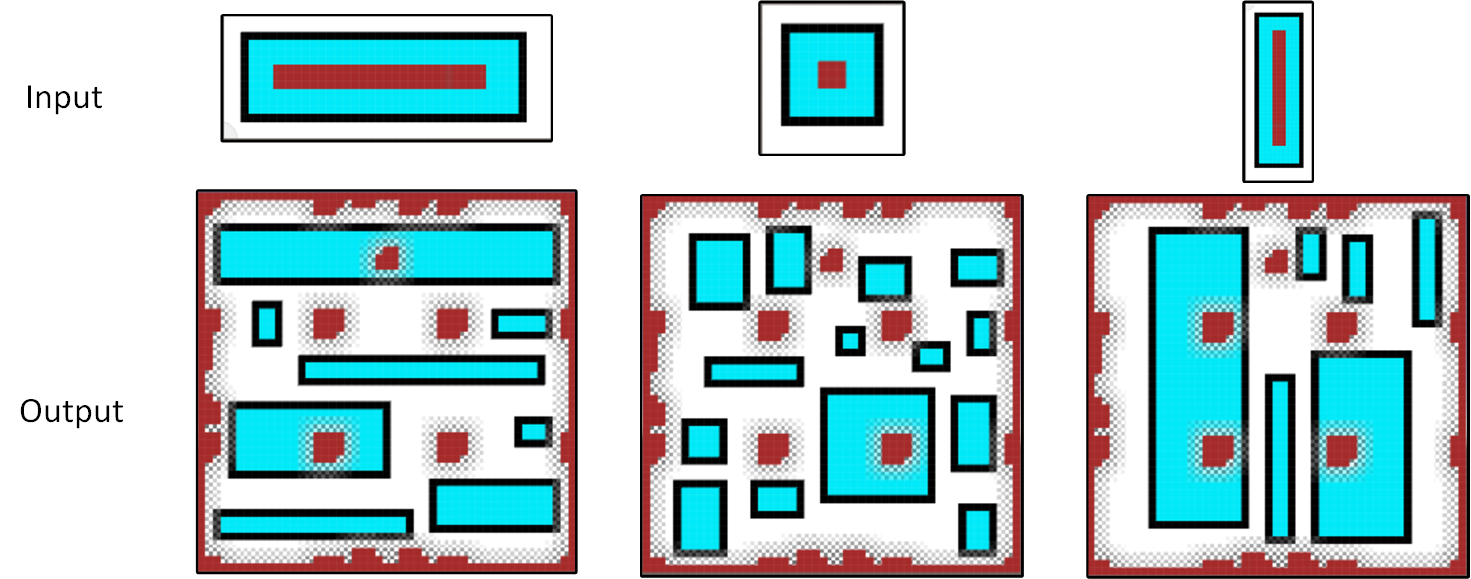}
    \caption{\begin{small}Comparison between output generated with different types of input. Rotations and reflections are disabled in order to accentuate the differences.\end{small}}
    \label{fig:RectangleInput}
\end{figure}

Using weights to select patterns also gives us control over the outcome of the algorithm.  We use the same pattern selection mechanism as in the base WFC algorithm: the frequency of each pattern extracted from the input gives a frequency distribution, which is then used to calculate the entropy of given tiles in order to select the next tile and pattern during the \textbf{observe} step.
The relative extent of identical features in the input thus gives control over the likelihood of choosing a corresponding pattern in the output.  Figure~\ref{fig:RectangleInput} shows the effect.  Even though horizontal and vertical rectangles and squares have identical sets of input patterns, the relative number of each pattern biases the output.

Further control is possible using the same mechanism. Tuning the ratio of white tiles to stretch tiles, for example, will affect the output in terms of how many paths there will be and how big paths will get. 

\subsubsection{Uniform distribution}

\begin{figure}[ht]
    \centering
    \includegraphics[width=2.8in]{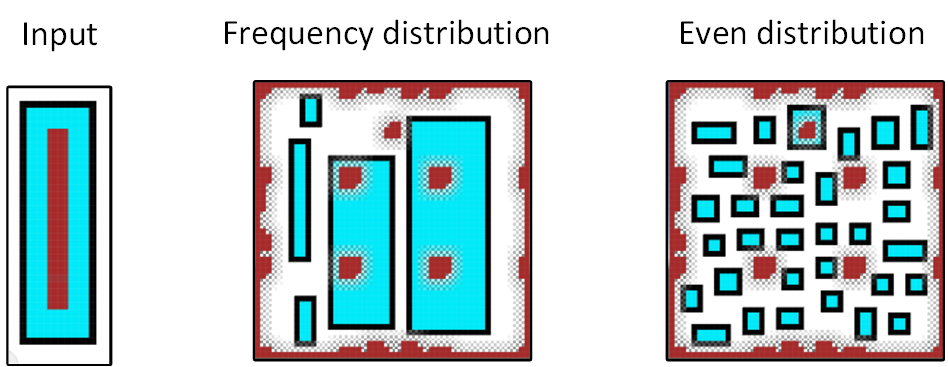}
    \caption{\begin{small}Comparison between output generated with different types of distribution. Rotations and reflections are disabled.\end{small}}
    \label{fig:RandomDistro}
\end{figure}

Pattern/tile selection can also be done without regard to pattern frequency. Figure~\ref{fig:RandomDistro} shows two outputs generated from the same input and random seed. The first output uses the normal frequency distribution, while the second output uses a uniform distribution. As we can see, the preponderance of vertical rectangles apparent in the first case is not reproduced in the second output, since the distribution of patterns is the same for each pattern. There is also less overall coverage, due to the fact that stretch space tiles are now less frequent than in the first output.

\subsubsection{Different input/output combinations}

To support the argument that the algorithm works for different combinations of input and output, we show examples on 2 larger inputs, namely \textit{Arena2} ($281 \times 209$) and \textit{Lak519d} ($168 \times 145$) from the \textit{Dragon Age: Origins} benchmark set.

\begin{figure}[ht]
    \centering
    \includegraphics[width=3.2in]{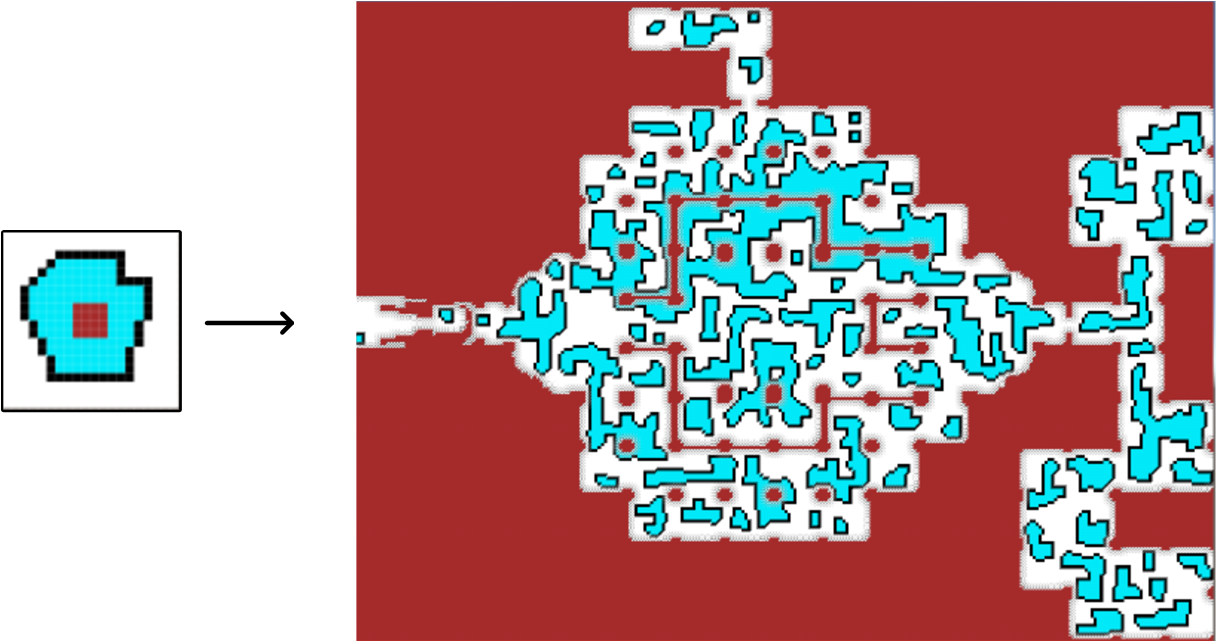}
    \caption{\begin{small}Example of execution with the Arena2 map.\end{small}}
    \label{fig:Arena2Example}
\end{figure}

Figure~\ref{fig:Arena2Example} shows a result of the algorithm on the \textit{Arena2} map.  Here we used a complex polygon as the input, incorporating both a concave notch and edges with different angles.  Rotations are reflections are also enabled.  The resulting paths are more organic, with many more complex turns and routes. 

\begin{figure}[ht]
    \centering
    \includegraphics[width=3in]{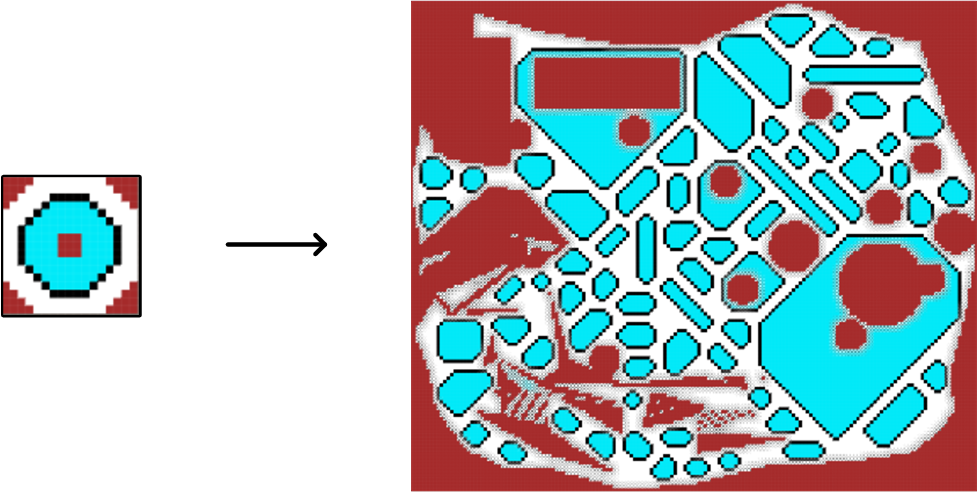}
    \caption{\begin{small}Example of execution with the Lak519d map.\end{small}}
    \label{fig:Lak519dExample}
\end{figure}

Figure~\ref{fig:Lak519dExample} shows a result on the \textit{Lak519d} map using an octagon as the input. For this input, enabling rotations and reflections is irrelevant as it would add no additional pattern. The more limited pattern set available in this output results in output quite different from figure~\ref{fig:Arena2Example}, with path shapes that have a strong similarity to the input.  Note that here we also added obstacle tiles in the input's corners to cover up white space and even out the ratio between white space and stretch space.

\subsection{Performance}
The performance of WFC mainly depends on 2 factors, the number of patterns extracted from the input, as well as the size of the generated output.  Our design integrates optimizations introduced by \citeauthor{FehrCourantFast} \shortcite{FehrCourantFast}.  We also make use of Unity's \textit{Job System}, using it to help parallelize the post-processing path generation step.

Measurements are gathered by running the algorithm 10 times on each test set and compare the change in average execution time. We discard the first execution for each test as a cache warm up.

\begin{table}[ht]
\begin{small}
\centering
\caption{\begin{small}Average time in seconds for varying inputs/outputs.\end{small}}
\label{table:inputoutput}
\centerline{
\begin{tabularx}{\columnwidth}{ |c|c|X|c|X|X| }
 \toprule
 Input & Output & Nb. Patt. & Size & Time (sec.) & Std Dev. \\ 
 \midrule
 Poly 2 & arena1 & 206 & $49\times49$ & 0.89 & 0.012 \\ 
 Poly 2 & orz000d & 253 & $79\times137$ & 7.61 & 0.118 \\ 
 Square & lak519d & 304 & $168\times145$ & 49.59 & 0.821 \\
 Octagon & lak519d & 320 & $168\times145$ & 53.11 & 1.024 \\ 
 Poly 1 & lak519d & 348 & $168\times145$ & 58.71 & 0.436 \\ 
 Poly 2 & lak519d & 376 & $168\times145$ & 61.02 & 1.388 \\
 Poly 2 & arena2 & 267 & $281\times209$ & 78.43 & 0.431 \\ 
 \bottomrule
\end{tabularx}
}
\end{small}
\end{table}

Table \ref{table:inputoutput} shows executions with different inputs and outputs. We used a specific input (Poly 2) with varying outputs as well as a specific Output (lak519d) with varying inputs. As shown by the average execution time (column Time), using inputs that generate more patterns increases the running time, however execution time grows notably more as the output size grows. Despite these variations, execution on significantly large maps is done in a reasonable amount of time.

\section{Related work}

Our work relies heavily on the WFC algorithm as it is an adaptation of the WFC overlap model. The WFC algorithm has seen a few notable ports and forks. \citeauthor{KarthWFC} released a paper on WFC in which they describe the algorithm in depth, discuss its influence and contextualize it in terms of constraint solving \cite{KarthWFC}. 
\citeauthor{FehrCourantFast} have developed a faster version of the algorithm written in C++ \cite{FehrCourantFast} which introduces modifications such as reducing the number of overlaps per pattern processed by the algorithm. 
\citeauthor{lefebvre:hal-01706539} have experimented with hiding messages in the generated images from the tiled model, with an emphasis on the fact that embedded images are not distinguishable from normal generated images \cite{lefebvre:hal-01706539}. 
Among a few others, \citeauthor{Gumin3dWFC} also developed a 3D version of the tiled model to generate 3D images using voxels \cite{Gumin3dWFC}.

\subsection{PCG}
Path generation itself is generically part of the field of \textit{Procedural Content Generation (PCG)}. This field applies mostly to games and consists of algorithms that generate content such as textures, levels, terrains, quests, items, or other game-related concepts. The key idea is that the content is created in a random and algorithmic way with minimal user input, in contrast to content being manually crafted by game designers.  Many generative techniques have been explored, including constraint models, based on understanding the rhythm or pacing of player experience, such as the \citeauthor{smith2010tanagra}'s \textit{Tanagra} tool for 2D platformer levels \shortcite{smith2010tanagra}, and machine-learning approaches that demonstrate it is possible to construct levels without domain-specific knowledge~\cite{snodgrass2018markov}. Although less common, PCG algorithms have been applied to path generation previously as well. \citeauthor{xu-14-generative} \shortcite{xu-14-generative} developed a tool to randomly place guards and cameras in stealth game levels.  Their approach concentrated on visibility and coverage properties, aiming to heuristically control level difficulty. More information on PCG can be found in \citeauthor{shaker2016procedural}'s book \shortcite{shaker2016procedural}.

\subsection{Roadmap Generation}
Higher-level navigation systems, such as roadmaps, can also be used as a basis for creating randomly wandering NPC motion, presenting seemingly goal-directed behaviour by defining random path choices at a larger scale. Paths based on optimal roadmaps, however, tend to scrape obstacles~\cite{nilsson69}, which is both not human-like, and a source of error in game simulation.  For games, NPC movement is better shifted away from obstacles.
Convex decompositions or \textit{navigation meshes} can be used to give more movement freedom.  Triangulations are a natural choice in this respect, and many systems build roadmaps from the triangulation dual.  \citeauthor{Demyen:2006:ETP:1597538.1597687} \shortcite{Demyen:2006:ETP:1597538.1597687}, for instance, show that it is efficient to build a graph for pathfinding by using constrained Delaunay triangulations, a technique also used by \citeauthor{DBLP:journals/corr/LensB16} \shortcite{DBLP:journals/corr/LensB16} to build roadmaps with controlled distance from obstacles.  \citeauthor{AmatoRandomizedRoadmap} \shortcite{AmatoRandomizedRoadmap} generate roadmaps by randomly selecting candidate points on the boundaries of a shape and using the midpoint between lines connecting pairs of candidates to form a roadmap.

In defining routes that well avoid obstacles, roadmap algorithms tend to approximate the \textit{medial axis skeleton} \cite{Blum:1967:ATF}, a simplified graph that defines points equally distant from internal obstacles. This has a natural appeal for roadmaps, and thus pathing, and use of the medial axis has been experimented with by many authors \cite{AmatoPRMMedialAxis,GuibasPRMMedialAxis}. \citeauthor{singh-15-using-TH} \shortcite{singh-15-using-TH} has taken this concept even further and used medial skeletons in a 3D environment (2D map with 3rd dimension as time) to generate roadmaps in a dynamic environment for stealth games.

Other pathfinding systems also lend themselves to path generation.
The hierarchical pathfinding algorithm developed by \citeauthor{botea2004near} \shortcite{botea2004near} creates an abstract graph of a map by separating the map into clusters, which could be adapted to be used generatively. 
Search graphs of various forms can be used in a similar fashion.
\textit{Rapidly exploring random trees (RRTs)}, for example, can be used to generate random paths in fixed maps \cite{LavalleRRT}. The idea of RRTs is to grow a randomized tree structure from a certain position in order to fill the map with many different random paths. This could be adapted to generate random meandering paths by setting a starting point and letting the RRT search grow, perhaps also including constraints on how it connects new nodes for controlling variety.

In contrast to pathfinding, dynamic systems can also be used to randomly move agents around obstacles. One can use steering behaviours \cite{reynolds1999steering} along with obstacle attraction \cite{fajen2003dynamical} to make an agent move around an obstacle. In a similar fashion, using flow fields for crowd pathing \cite{emerson2013crowd}, one can generate flow fields that gravitate around obstacles and make agents move along those flow fields.

\section{Conclusion \& Future Work}

In this paper, we have presented a modification to the Wave Function Collapse algorithm that handles random path generation using input texture samples and fixed output maps. The algorithm extends the original WFC algorithm by generating paths similar to the input by using stretch space, and by using masks to enable new patterns based on the shape of obstacles from the output map. We have included post-processing options to filter out small paths and to make remaining paths smoother. We presented a convenient tool developed in Unity that lets any user, with our without programming knowledge, use the algorithm for path generation purposes.

Using a texture generation algorithm to generate paths has a few inherent limitations.
First of all, there is no defined way to enforce true global constraints on the output to satisfy the global shape of the input path.
Moreover, since paths are defined by pixels, intersecting paths introduce ambiguity in converting the 2D image into waypoints for a game level. 

Future work on the algorithm could include using heuristics on intersecting tiles to determine which neighbour is the next path segment. Further development include enforcing additional constraints to make paths meet certain evaluation criteria, such as the number of turns taken or the number of obstacles that paths enclose. A more complex constraint would be to augment our algorithm with a frequency distribution on pairs of patterns, which could bind the selection of a pattern to previously selected neighbour patterns. 

\clearpage

\bibliographystyle{aaai}
\bibliography{main}

\begin{thebibliography}{}

\bibitem[\protect\citeauthoryear{Amato and Wu}{1996}]{AmatoRandomizedRoadmap}
Amato, N.~M., and Wu, Y.
\newblock 1996.
\newblock A randomized roadmap method for path and manipulation planning.
\newblock In {\em Proceedings of IEEE International Conference on Robotics and
  Automation}, volume~1,  113--120 vol.1.

\bibitem[\protect\citeauthoryear{Blum}{1967}]{Blum:1967:ATF}
Blum, H.
\newblock 1967.
\newblock A transformation for extracting new descriptors of shape.
\newblock In Wathen-Dunn, W., ed., {\em Models for the Perception of Speech and
  Visual Form}. Cambridge: MIT Press.
\newblock  362--380.

\bibitem[\protect\citeauthoryear{Botea, M{\"u}ller, and
  Schaeffer}{2004}]{botea2004near}
Botea, A.; M{\"u}ller, M.; and Schaeffer, J.
\newblock 2004.
\newblock Near optimal hierarchical path-finding.
\newblock {\em Journal of game development} 1(1):7--28.

\bibitem[\protect\citeauthoryear{Chaikin}{1974}]{Chaikin1974}
Chaikin, G.~M.
\newblock 1974.
\newblock An algorithm for high-speed curve generation.
\newblock {\em Computer Graphics and Image Processing} 3(4):346 -- 349.

\bibitem[\protect\citeauthoryear{Demyen and
  Buro}{2006}]{Demyen:2006:ETP:1597538.1597687}
Demyen, D., and Buro, M.
\newblock 2006.
\newblock Efficient triangulation-based pathfinding.
\newblock In {\em Proceedings of the 21st National Conference on Artificial
  Intelligence - Volume 1}, AAAI'06,  942--947.
\newblock AAAI Press.

\bibitem[\protect\citeauthoryear{Douglas and Peucker}{1973}]{DouglasPeucker}
Douglas, D.~H., and Peucker, T.~K.
\newblock 1973.
\newblock Algorithms for the reduction of the number of points required to
  represent a digitized line or its caricature.
\newblock {\em Cartographica: The International Journal for Geographic
  Information and Geovisualization} 10(2):112 -- 122.

\bibitem[\protect\citeauthoryear{Emerson}{2013}]{emerson2013crowd}
Emerson, E.
\newblock 2013.
\newblock Crowd pathfinding and steering using flow field tiles.
\newblock {\em Game AI Pro: Collected Wisdom of Game AI Professionals}
  307--316.

\bibitem[\protect\citeauthoryear{Fajen \bgroup et al\mbox.\egroup
  }{2003}]{fajen2003dynamical}
Fajen, B.~R.; Warren, W.~H.; Temizer, S.; and Kaelbling, L.~P.
\newblock 2003.
\newblock A dynamical model of visually-guided steering, obstacle avoidance,
  and route selection.
\newblock {\em International Journal of Computer Vision} 54(1-3):13--34.

\bibitem[\protect\citeauthoryear{Fehr and Courant}{2018}]{FehrCourantFast}
Fehr, M., and Courant, N.
\newblock 2018.
\newblock fast-wfc.
\newblock \url{https://github.com/math-fehr/fast-wfc}.
\newblock Github repository.

\bibitem[\protect\citeauthoryear{Guibas, Holleman, and
  Kavraki}{1999}]{GuibasPRMMedialAxis}
Guibas, L.~J.; Holleman, C.; and Kavraki, L.~E.
\newblock 1999.
\newblock A probabilistic roadmap planner for flexible objects with a workspace
  medial-axis-based sampling approach.
\newblock In {\em Proceedings 1999 IEEE/RSJ International Conference on
  Intelligent Robots and Systems: Human and Environment Friendly Robots with
  High Intelligence and Emotional Quotients}, volume~1,  254--259.

\bibitem[\protect\citeauthoryear{Gumin}{2016a}]{Gumin3dWFC}
Gumin, M.
\newblock 2016a.
\newblock {Basic3DWFC}.
\newblock \url{https://bitbucket.org/mxgmn/basic3dwfc/src/master/}.
\newblock Bitbucket repository.

\bibitem[\protect\citeauthoryear{Gumin}{2016b}]{GuminWFC}
Gumin, M.
\newblock 2016b.
\newblock {W}ave{F}unction{C}ollapse.
\newblock \url{https://github.com/mxgmn/WaveFunctionCollapse}.
\newblock Github repository.

\bibitem[\protect\citeauthoryear{Karth and Smith}{2017}]{KarthWFC}
Karth, I., and Smith, A.~M.
\newblock 2017.
\newblock {W}ave{F}unction{C}ollapse is constraint solving in the wild.
\newblock In {\em Proceedings of the 12th International Conference on the
  Foundations of Digital Games}, FDG '17,  68:1--68:10.
\newblock New York, NY, USA: ACM.

\bibitem[\protect\citeauthoryear{Lavalle}{1998}]{LavalleRRT}
Lavalle, S.~M.
\newblock 1998.
\newblock Rapidly-exploring random trees: A new tool for path planning.
\newblock Technical Report TR 98-11, Iowa State University.

\bibitem[\protect\citeauthoryear{Lawler}{1991}]{lawler}
Lawler, G.~F.
\newblock 1991.
\newblock {\em Intersections of random walks}.
\newblock Birkh{\"a}user.

\bibitem[\protect\citeauthoryear{Lefebvre, Wei, and
  Barnes}{2018}]{lefebvre:hal-01706539}
Lefebvre, S.; Wei, L.-Y.; and Barnes, C.
\newblock 2018.
\newblock Informational texture synthesis.
\newblock \url{https://hal.inria.fr/hal-01706539/file/infotexsyn.pdf}.
\newblock working paper or preprint.

\bibitem[\protect\citeauthoryear{Lens and
  Boigelot}{2016}]{DBLP:journals/corr/LensB16}
Lens, S., and Boigelot, B.
\newblock 2016.
\newblock From constrained {D}elaunay triangulations to roadmap graphs with
  arbitrary clearance.
\newblock {\em CoRR} abs/1606.02055.
\newblock \url{http://arxiv.org/abs/1606.02055}.

\bibitem[\protect\citeauthoryear{Nilsson}{1969}]{nilsson69}
Nilsson, N.~J.
\newblock 1969.
\newblock A mobile automaton: An application of artificial intelligence
  techniques.
\newblock In {\em 1st International Conference on Artificial Intelligence},
  509--520.

\bibitem[\protect\citeauthoryear{Ramer}{1972}]{RAMER1972244}
Ramer, U.
\newblock 1972.
\newblock An iterative procedure for the polygonal approximation of plane
  curves.
\newblock {\em Computer Graphics and Image Processing} 1(3):244 -- 256.

\bibitem[\protect\citeauthoryear{Reynolds}{1999}]{reynolds1999steering}
Reynolds, C.~W.
\newblock 1999.
\newblock Steering behaviors for autonomous characters.
\newblock In {\em Game developers conference}, volume 1999,  763--782.

\bibitem[\protect\citeauthoryear{Shaker, Togelius, and
  Nelson}{2016}]{shaker2016procedural}
Shaker, N.; Togelius, J.; and Nelson, M.~J.
\newblock 2016.
\newblock {\em Procedural Content Generation in Games: A Textbook and an
  Overview of Current Research}.
\newblock Springer.

\bibitem[\protect\citeauthoryear{Singh}{2015}]{singh-15-using-TH}
Singh, D.
\newblock 2015.
\newblock Using medial skeleton for path finding in dynamic stealth games.
\newblock Master's thesis, McGill University, Montr{\'e}al, Canada.

\bibitem[\protect\citeauthoryear{Smith, Whitehead, and
  Mateas}{2010}]{smith2010tanagra}
Smith, G.; Whitehead, J.; and Mateas, M.
\newblock 2010.
\newblock Tanagra: A mixed-initiative level design tool.
\newblock In {\em Proceedings of the Fifth International Conference on the
  Foundations of Digital Games},  209--216.
\newblock ACM.

\bibitem[\protect\citeauthoryear{Snodgrass}{2018}]{snodgrass2018markov}
Snodgrass, S.
\newblock 2018.
\newblock {\em Markov Models for Procedural Content Generation}.
\newblock Ph.D. Dissertation, Drexel University.

\bibitem[\protect\citeauthoryear{Sturtevant}{2012}]{sturtevant2012benchmarks}
Sturtevant, N.
\newblock 2012.
\newblock Benchmarks for grid-based pathfinding.
\newblock {\em Transactions on Computational Intelligence and AI in Games}
  4(2):144 -- 148.

\bibitem[\protect\citeauthoryear{Wilmarth, Amato, and
  Stiller}{1999}]{AmatoPRMMedialAxis}
Wilmarth, S.~A.; Amato, N.~M.; and Stiller, P.~F.
\newblock 1999.
\newblock {MAPRM}: a probabilistic roadmap planner with sampling on the medial
  axis of the free space.
\newblock In {\em Proceedings 1999 IEEE International Conference on Robotics
  and Automation}, volume~2,  1024--1031 vol.2.

\bibitem[\protect\citeauthoryear{Xu, Tremblay, and
  Verbrugge}{2014}]{xu-14-generative}
Xu, Q.; Tremblay, J.; and Verbrugge, C.
\newblock 2014.
\newblock Generative methods for guard and camera placement in stealth games.
\newblock In {\em Tenth Annual AAAI Conference on Artificial Intelligence and
  Interactive Digital Entertainment (AIIDE 2014)}.

\end{thebibliography}

\end{document}